\documentclass[sigconf,natbib=true,anonymous=false]{acmart}

\makeatletter
\def\@ACM@checkaffil{
    \if@ACM@instpresent\else
    \ClassWarningNoLine{\@classname}{No institution present for an affiliation}%
    \fi
    \if@ACM@citypresent\else
    \ClassWarningNoLine{\@classname}{No city present for an affiliation}%
    \fi
    \if@ACM@countrypresent\else
        \ClassWarningNoLine{\@classname}{No country present for an affiliation}%
    \fi
}
\makeatother

\def\CHK#1 {\textcolor{magenta}{{\bf [CHK:}~#1{\bf ]}}~}
\def\ADD#1 {\textcolor{cyan}{{\bf [ADD:}~#1{\bf}]}~}

\usepackage[utf8]{inputenc}

\usepackage{pifont} 
\newcommand{\cmark}{\ding{51}}%
\newcommand{\xmark}{\ding{55}}%

\usepackage[normalem]{ulem}
\useunder{\uline}{\ul}{} 

\usepackage{subcaption} 

\usepackage{multirow} 

\usepackage{amsmath,empheq} 

\usepackage[ruled,linesnumbered]{algorithm2e} 
\SetKwComment{Comment}{/* }{ */} 

\usepackage{enumitem}
\setlist[itemize]{leftmargin=*}
\setlist[enumerate]{leftmargin=*}

\usepackage[title]{appendix}

\renewcommand{\figureautorefname}{Figure}

\def\equationautorefname~#1\null{Eqn. ~(#1)\null}
\def\figureautorefname~#1\null{Fig. ~#1\null}

\newtheoremstyle{remboldstyle}
  {}{}{\itshape}{}{\bfseries}{:}{.5em}{{\thmname{#1 }}{\thmnumber{#2.}}{\thmnote{ #3}}}
\theoremstyle{remboldstyle}

\acmConference[MM'23]{ACM MM'23: ACM International Conference on Multimedia}{October 29-November 3, 2023}{Ottawa, Canada}
\acmBooktitle{Proceedings of the 30th ACM International Conference on Multimedia (MM '23), October 29-November 3, 2023, Ottawa, Canada}

\begin{document}

\begingroup
\hyphenpenalty 9900

\title{RTQ: Rethinking Video-language Understanding Based on Image-text Model}

\author{Xiao Wang}
\email{scz.wangxiao@gmail.com}
\affiliation{%
  \institution{Harbin Institute of Technology, Shenzhen \& JD.com Inc.}
}

\author{Yaoyu Li}
\email{liyaoyu2014@gmail.com}
\affiliation{%
  \institution{JD.com Inc.}
}

\author{Tian Gan}
\email{gantian@sdu.edu.cn}
\authornote{Corresponding author: Tian Gan}
\affiliation{%
  \institution{Shandong University}
}

\author{Zheng Zhang}
\email{zhangzheng11@jd.com}
\affiliation{%
  \institution{JD.com Inc.}
}

\author{Jingjing Lv}
\email{lvjinghit@163.com}
\affiliation{%
  \institution{JD.com Inc.}
}

\author{Liqiang Nie}
\email{nieliqiang@gmail.com}
\affiliation{%
  \institution{Harbin Institute of Technology, Shenzhen}
}

\renewcommand{\shortauthors}{Xiao Wang et al.}

\begin{abstract}
    Recent advancements in video-language understanding have been established on the foundation of image-text models, resulting in promising outcomes due to the shared knowledge between images and videos. 
    However, video-language understanding presents unique challenges due to the inclusion of highly complex semantic details, which result in \textit{information redundancy}, \textit{temporal dependency}, and \textit{scene complexity}.
    Current techniques have only partially tackled these issues, and our quantitative analysis indicates that some of these methods are complementary.
    In light of this, we propose a novel framework called RTQ (Refine, Temporal model, and Query), 
    which addresses these challenges simultaneously.
    The approach involves refining redundant information within frames, modeling temporal relations among frames, and querying task-specific information from the videos.
    Remarkably, our model demonstrates outstanding performance even in the absence of video-language pre-training, and the results are comparable with or superior to those achieved by state-of-the-art pre-training methods.
    
\end{abstract}




\maketitle

\section{introduction}

Video-language understanding ability can reflect the proficiency of intelligent agents in perceiving and interpreting visual and textual cues in the real world. 
This ability is evaluated through a range of tasks, including text-to-video retrieval, video captioning, and video question answering, etc. 
Recent approaches in this area generally modify pre-trained image-text models for video-language understanding, owing to the transferable knowledge acquired by these models \cite{lei_clipbert_2021, xue_clip-vip_2023, GAN_CVPR_2023}. 
Since image-text models have learned a lot of transferable vision-language knowledge, these approaches generally perform better and are becoming the \textit{de facto} paradigm.
However, such an approach has limitations in handling situations beyond the shared knowledge between images and videos.

\begin{figure}[t]
    \centering
    \includegraphics[width=\linewidth]{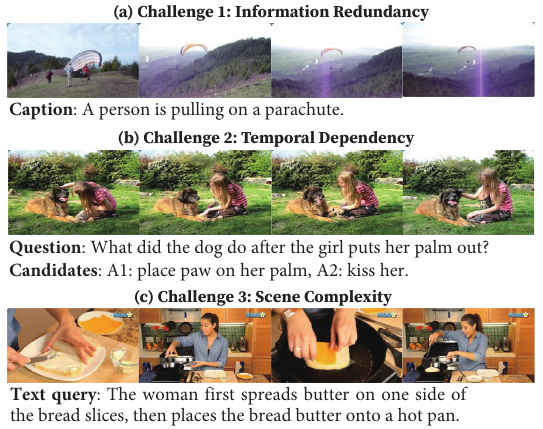}
    \vspace{-2em}
    \caption{Video language understanding requires dynamic perception and interpretation of complex semantics, which can be further decomposed into three challenges.}
    \vspace{-1em}
    \label{fig:intro_figure}
\end{figure}

Video-language understanding involves dynamic perception and interpretation of complex semantics.
This can be further decomposed into three challenges as depicted in \autoref{fig:intro_figure}.
The first challenge is \textbf{Information redundancy} which arises due to the presence of information that is duplicated or lacks semantic meaning. It hinders the model's ability to accurately recognize essential cues. 
For example, frames in \autoref{fig:intro_figure}(a) are quite similar, some of which can be removed without affecting the interpretation.
The second challenge is \textbf{Temporal dependency} which requires the model to identify and understand the relationships between video frames. 
Consider video in \autoref{fig:intro_figure}(b), multiple actions occur between the girl and the dog. To correctly answer the given question, the model has to recognize the order of these actions.
The third challenge is \textbf{Scene complexity}, where a video depicts multiple concepts, but only some of them are task-relevant.
In such cases, the model needs to prioritize task-relevant information to achieve better performance. If a model fails to consider this, it may become overwhelmed by the complexity of the scene, leading to poor performance.
For instance, in video of \autoref{fig:intro_figure}(c), numerous cooking actions and ingredients are presented. 
To achieve better performance, the model must concentrate on crucial pieces of information such as objects and actions mentioned in text queries, captions, or questions.


Current approaches have mainly focused on addressing one or two of the aforementioned challenges. 
However, it is essential to consider these challenges jointly since they address different facets that  could complement each other. 
For example, eliminating redundant information could benefit the model when solving the scene complexity problem \cite{Dong_CVPR_SHA}. 
Nonetheless, it is crucial to carry out quantitative analysis to ascertain the extent of the complementarity between these challenges.
Moreover, joint modeling is not a straightforward task due to the challenge of cooperative design \cite{Wang_TMM_2022,Gan_tomccap_19}. 
Existing methods \cite{zhao_centerclip_2022, liu_ts2-net_2022} have addressed the information redundancy challenge by selecting meaningful tokens or frames. 
However, this selection process breaks the spatial consistency between frames, making it intractable for temporal modeling techniques. 
In this regard, TS2-Net \cite{liu_ts2-net_2022} set the selection module as the final layer to enable temporal modeling. Nevertheless, this approach is sub-optimal since a lot of redundant information could be reduced in the shallower layers. 
On the other hand, some methods choose to ignore the information redundancy challenge, and only address the temporal dependency and scene complexity problems \cite{wang_omnivl_2022, xu_mplug-2_2023}.

This paper begins with a quantitative analysis of existing  approaches, wherein we cluster them based on their predictions. Our analysis reveals that the models within each cluster address the same challenges and confirms their complementarity, as previously discussed.
Based on this observation, we propose the \textbf{R}efine, \textbf{T}emporal model, and \textbf{Q}uery (\textbf{RTQ}) framework to jointly tackle the aforementioned challenges. The RTQ framework consists of three key components, each of which is designed to address a specific challenge. 
The first refinement component eliminates redundant patches across adjacent video frames using clustering, followed by the selection of representative patches.
The second temporal modeling component uses an image backbone augmented with a temporal module. The module is designed to perceive and interpret temporal patterns in the video, without requiring spatial consistency between patches across frames.
In the third query component, we adopt language query (text query, question, or generated caption) to accumulate task-relevant information gradually. 
The aforementioned three components can be realized through any appropriate methods. In this study, we select the simplest modules available. Specifically, we employ the non-parametric \textit{k-medoids++} method for clustering, the message token mechanism for the temporal module, and cross-attention for the query. Despite their simplicity and the absence of pretraining, our approach achieves superior (or comparable) performance to the pre-training based methods in text-to-video retrieval, video captioning, and video question answering.

In summary, our contributions are threefold:
\begin{itemize}
    \item Our systemic analysis reveals that current methods focus only on restricted aspects of video-language understanding, and they are complementary.
    \item We propose the RTQ framework to jointly model information redundancy, temporal dependency, and scene complexity in video-language understanding.
    \item We demonstrate that, even without pre-training on video-language data, our method can achieve superior (or comparable) performance with state-of-the-art pre-training methods. We will make our code publicly available for further research\footnote{See our GitHub repository \href{https://github.com/SCZwangxiao/RTQ-MM2023}{https://github.com/SCZwangxiao/RTQ-MM2023}.}.
\end{itemize}

\section{related works} \label{sec:related_works}

\subsection{Video-language Pre-training}

The dominant pre-training methods for video-language understanding fall into two categories. 
The first category 
focus on data curation and refinement. 
For example, 
Zellers \textit{et al.} \cite{zellers_merlot_2021} collect a  diverse corpus of frames/ASR, named YT-Temporal-180M, from videos covering authentic situations, which improves downstream performance compared to curated instructional video corpora. 
However, 
video-language pre-training also suffers from incoherence and misalignment between ASR/subtitle and video \cite{GAN_CVPR_2023}. 
To overcome this issue, 
Bain \textit{et al.} \cite{bain_frozen_2021} curate a video dataset, WebVid, with well-aligned textual description annotations.
CLIP-ViP \cite{xue_clip-vip_2023} uses an image captioning model to generate captions for the middle frame of videos in HD-VILA-100M to obtain more aligned video-text annotations.
The texts are not so well aligned compared with videos in WebVid, but the scale of data is much bigger.

The second category 
focus on improving pre-training strategy.
To bridge the modality gap between video and text, BridgeFormer \cite{ge_bridgeformer_2022} proposed a novel multiple-choice questions task to achieve fine-grained video-text interactions.
OmniVL \cite{wang_omnivl_2022}  and mPLUG-2 \cite{xu_mplug-2_2023} explore a universal paradigm that benefit from joint modality learning. 
OmniVL adopts a unified transformer-based visual encoder for both image and video inputs, facilitating joint image-language and video-language pre-training.
On the other hand, mPLUG-2 introduces a multi-module composition network that
shares common universal modules for modality collaboration and disentangles  different modality modules to handle modality entanglement.

Despite their contributions, all of the above methods mainly focus on pre-training data or strategy, neglecting architecture design for modeling information redundancy, temporal dependency, and scene complexity in video-language understanding, which prevents the full realization of the model's potential.


\subsection{Video-language Model Architecture} \label{sec:related_arch}
Classical architectures utilize separately pre-trained vision and language backbones, which remain static during training \cite{wei2019neural}. 
However, recent studies have identified limitations in these approaches related to modality and domain gaps~\cite{lei_clipbert_2021},
whereas newer architectures based on image-text pre-training models show more promising results due to their ability to bridge the modality gap and enhance transferability \cite{radford_clip_2021}. 
As videos are essentially composed of image sequences, the insights from image-text models can also be applied to video-language understanding.

Recent architectures can be categorized into three groups: refinement, temporal modeling, and information query. 
\textbf{Refinement methods} aim to identify and eliminate irrelevant image patches or frames. In the case of TS2-Net \cite{liu_ts2-net_2022}, a scoring network is initially employed to assess the significance of each patch in the frames, followed by the application of a differentiable TopK function for patch selection. Zhao \textit{et al.} \cite{zhao_centerclip_2022} proposed to first cluster all patches in adjacent frames and then retain only the patches that are closest to each cluster centroid. 
\textbf{Temporal modeling methods} focus on modeling the temporal dependencies between video frames. Frozen \cite{bain_frozen_2021} and STAN \cite{liu_stan_2023} insert a temporal attention layer in their image model. Their difference is that STAN \cite{liu_stan_2023} inserts the layer in parallel instead of sequential with the original model, which empirically performs better. X-CLIP \cite{ni_x-clip_2022} and CLIP-ViP \cite{xue_clip-vip_2023} adopt the message tokens mechanism, where each frame has message token(s) to communicate with other frames. They differ in the construction approach and perception range of message tokens. 
CLIP4clip \cite{luo_clip4clip_2022} employs a transformer to model temporal dependencies, while TS2-Net shifts spatial token features across adjacent frames to capture local movement. 
%
\textbf{Information query methods} are capable of mining task-relevant information from the whole video.
Gorti \textit{et al.} \cite{gorti_x-pool_2022} proposed to use text as the query to guide the aggregation of useful information among the entire video. VideoCoCa \cite{yan_videococa_2023} leverages a generative pooling mechanism that gradually accumulates relevant information for caption generation.

However, these methods only address one or two aspects of video understanding and fail to consider the complementarity of different architectures.



\section{Preliminary Analysis} \label{sec:analysis}

\begin{figure}[t]
    \centering
    \includegraphics[width=0.99\linewidth]{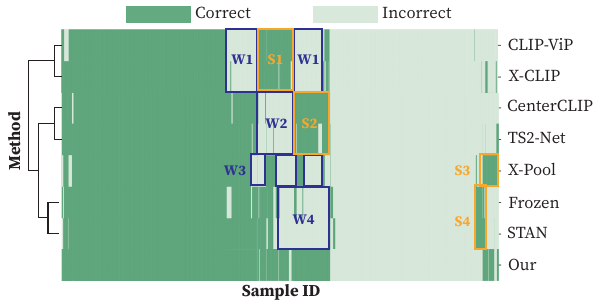}
    \vspace{-1em}
    \caption{Clustering results of existing methods in the NeXt-QA \cite{xiao2021next} dataset.}
    \vspace{-1em}
    \label{fig:cluster_results}
\end{figure}

In \autoref{sec:related_arch}, we highlighted that existing video-language models do not simultaneously address all three challenges and that we believe a joint approach is beneficial as the challenges may be complementary. 
To gain more objective insights, we conducted a quantitative analysis by clustering the models based on their successful and failed cases. Our rationale was to examine their complementarity by comparing representative cases within each cluster. In the following sections, we explain our clustering methodology and present our findings.


\subsection{Clustering Method}
To conduct a clustering analysis, it is necessary to establish a vector representation of a model to serve as the basis for measuring the distance between models \cite{zhong2021graph}. 
Subsequently, the appropriate clustering methods are selected for analysis. 
 In this study, we define a method's representation as $\mathbf{m}\in\mathbb{R}^N$, where $N$ is the number of samples in the validation set, and $m_i\in\{0, 1\}$ indicates whether the $i$-th sample is correctly predicted. 
 Hamming distance $d(\mathbf{m}, \mathbf{n})$ is used to assess the similarity of two methods $\mathbf{m}$ and $\mathbf{n}$, given by:
\begin{equation}
    d(\mathbf{m}, \mathbf{n}) = \sum_{i=1}^N{\mathbb{I}(m_i=n_i)},
\end{equation}
where $\mathbb{I}(m_i=n_i)=1$ only if $m_i=n_i$. 
Finally, we employ hierarchical clustering to explore the relationships among each method.

Our analysis was conducted on the validation split of the NeXt-QA dataset~\cite{xiao2021next} to perform our analysis because it contains a considerable number of descriptional, temporal, and causal questions that thoroughly evaluate a model's video understanding capabilities~\cite{buch_atp_2022}.

\subsection{Result Analysis}
The clustering results are depicted in \autoref{fig:cluster_results}, with each row representing a distinct method and each column a unique sample ID (approximately  5k IDs).
The results revealed four clusters of methods, namely: 
(1) temporal modeling methods that leverage message tokens mechanism, such as CLIP-ViP \cite{xue_clip-vip_2023}, X-CLIP \cite{ni_x-clip_2022}; 
(2) refinement methods, including CenterClip \cite{zhao_centerclip_2022}, TS2-Net \cite{liu_ts2-net_2022}; 
(3) information query approach XPool \cite{gorti_x-pool_2022}; 
and (4) temporal modeling methods based on temporal attention, such as Frozen \cite{bain_frozen_2021} and STAN \cite{liu_stan_2023}. 
The four clusters broadly align with the three categories of methods discussed in \autoref{sec:related_works}, namely refinement, temporal modeling, and information query.

Upon closer examination of the results, we find that each cluster of methods has distinct advantages in handling certain types of samples that other methods may struggle with (as indicated by the Area S1-S4 in \autoref{fig:cluster_results}). 
Meanwhile, these methods also exhibit weaknesses in handling particular samples, but which can be overcome effectively by other methods (as indicated by the Area W1-W4). 
Our findings indicate that current approaches tend to focus on specific aspects of video-language understanding, but exhibit some degree of complementarity. 
Therefore, it is possible and beneficial to jointly consider all three challenges to harness the strengths of these methods while mitigating their respective weaknesses. 
Based on these insights, we propose the RTQ framework, which we elaborate in the next section.

\section{method}

\subsection{Overview}

Based on observations in \autoref{sec:analysis}, 
we propose addressing the three challenges of video-language understanding collaboratively in the RTQ framework, outlined in
\autoref{fig:overview} (a).
The framework leverages a video encoder with refinement and temporal modules, and a query component composed of a Mixture of Encoder-Decoder (MoED).

\begin{figure*}[t]
    \centering
    \includegraphics[width=0.99\linewidth]{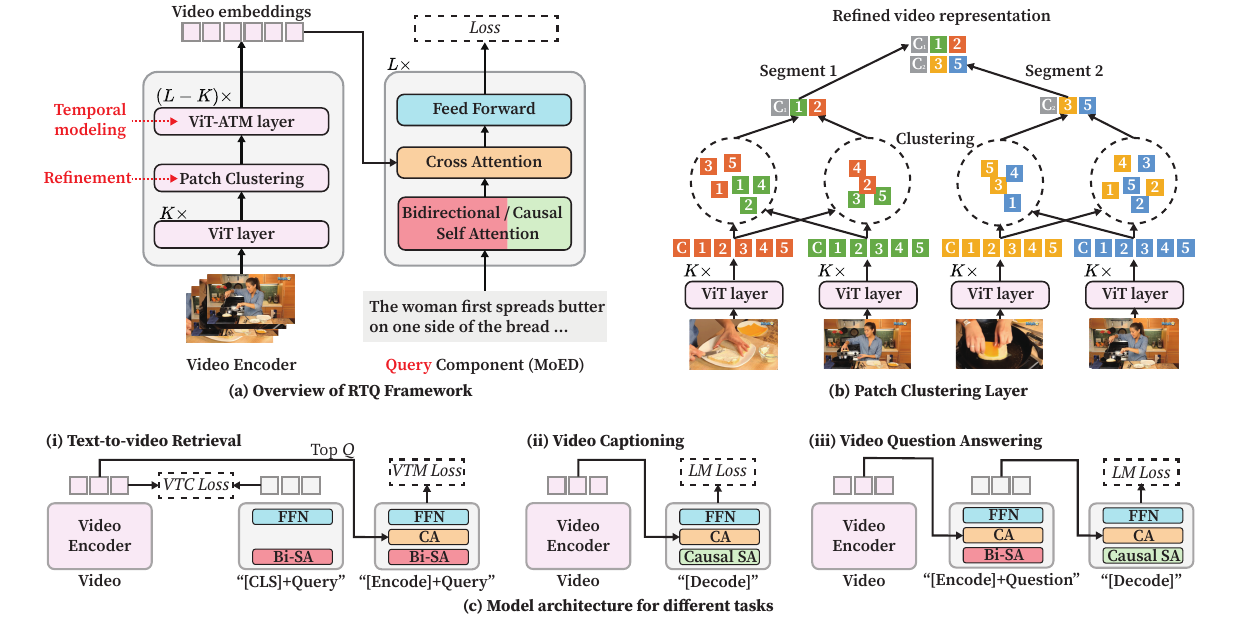}
    \vspace{-1em}
    \caption{Method overview.}
    \vspace{-1em}
    \label{fig:overview}
\end{figure*}

Specifically, the video encoder first encodes individual video frames utilizing $K$ Vision Transformer (ViT) \cite{dosovitskiy_vit_2021} layers, thereby generating image patch embeddings with semantic meanings. 
Then, a patch clustering layer is employed to eliminate redundant patches by retaining only representative patches from the clustering results. 
Subsequently, the remaining patches are fed into $(L-K)$ ViT Augmented with Temporal Module (ViT-ATM) layers, where $L$ is the total number of video encoder layers. These layers enable the model to capture temporal dependencies between frames and produce video embeddings as the output.
Finally, in the query component of the framework, task-specific text queries in each layer are employed to progressively collect task-relevant information from the video embeddings. 
This process varies depending on tasks, which we elaborate in \autoref{sec:query_component}.

\vspace{-1em}
\subsection{Refinement Component}
As illustrated in \autoref{fig:overview} (b), patch clustering layer is employed to refine the video embeddings. 
This process entails grouping patches in adjacent frames and selecting the representative ones to generate refined video embeddings. 

The output frame embeddings of the ViT layers are denoted as
$\mathbf{\hat{V}}^K\in\mathbb{R}^{F\times(1+P_{\textmd{in}})\times d}$,
where $F$ is the number of input frames, ``1'' represents the \texttt{[CLS]} token, $P_{\textmd{in}}$ is the number of patches per image divided by ViT, 
and $d$ is the hidden dimension. 
%
Initially, frames are grouped into $S$ segments,
with each containing $F/S$ frames.
Next, the \textit{k-medoids++} method \cite{zhao_centerclip_2022} is applied to cluster all $F/S(1+P_{\textmd{in}})$ patches within each segment, 
resulting in $(1+P_{\textmd{out}})$ clusters. 
%
Finally, $(1+P_{\textmd{out}})$ patches that are closest to each cluster centroid are selected to form the refined video embeddings
$\mathbf{V}^K\in\mathbb{R}^{S\times(1+P_{\textmd{out}})\times d}$.

It is worth noting that the clustering process is not limited to the \textit{k-medoids++} method, and may be implemented with other clustering methods. The straightforward and non-parametric \textit{k-medoids++} method is employed in this study to demonstrate the effectiveness of our proposed framework.

\vspace{-1em}
\subsection{Temporal Modeling Component} \label{sec:t_component}
The temporal modeling is accomplished by incorporating temporal modules into the ViT layers. The temporal module generally consists of a modified or inserted layer in the original ViT layer.
Temporal modules come in three varieties: 
temporal attention \cite{bain_frozen_2021}, 
message tokens \cite{xue_clip-vip_2023}, 
and temporal shifting \cite{liu_ts2-net_2022}. 
We adopt the message tokens mechanism, the details and design rationales are explained below.

Both temporal attention and temporal shifting mechanisms necessitate spatial consistency among patch positions of input frames. 
This is because they perform temporal reasoning on patches that are located at the same positions across different frames. 
However, clustering disrupts spatial consistency in our framework, as illustrated in~\autoref{fig:overview} (b).
Segment 1 has patches from location ID ``1, 2'', while  segment 2 has patches from location ID ``3, 5''. 
Furthermore, 
temporal reasoning is built upon the underlying assumption that content within each patch changes gradually, enabling the identification of temporal semantics by observing patches across different frames.
This assumption is not applicable in the video-language understanding task because it predominantly involves processing lengthy, untrimmed videos. 
Our ablation studies in \autoref{sec:ablation} have confirmed this view.

Different from these two mechanisms, the message token mechanism generally inserts several learnable embeddings along with patch embeddings, and utilize these learnable embeddings as an intermediary for temporal reasoning between frames. 
In our framework, we utilize a straightforward message token mechanism similar to that in X-CLIP \cite{ni_x-clip_2022}. 
Specifically, in the $k$-th layer, we first select and concatenate the \texttt{[CLS]} token in the refined video embedding to produce message token embeddings $\mathbf{\hat{m}}^k\in\mathbb{R}^{S\times d}$. 
We then perform Self-Attention (SA) along these tokens to learn the temporal dependencies between frames:
\begin{equation}
    \mathbf{m}^k = \mathbf{\hat{m}}^k + \textmd{SA} \left( \mathbf{\hat{m}}^k \right ).
\end{equation}

Finally, $\mathbf{m}^k$ and other patch tokens are fed into the original ViT layer. After $(L-K)$ ViT-ATM layer, our method produces the video embeddings video embeddings
$\mathbf{V}^L\in\mathbb{R}^{S\times(1+P_{\textmd{out}})\times d}$.


\subsection{Query Component} \label{sec:query_component}

After the previous refinement and temporal modeling components, videos have been encoded into a temporal-aware representation with high information density. Nevertheless, a significant amount of task-irrelevant information remains due to the challenge of scene complexity. 
For instance, in retrieval and question answering tasks, the model should focus on information about the query or question. Similarly, in captioning, the model should emphasize previously undescribed objects and events in the current generated caption.

To address this issue, we introduce a query component that regards video embeddings as memories and use task-specific queries to gradually gather relevant details for the final results. 
To reuse common structures among various tasks, we integrate an $L$ layer MoED \cite{li_blip_2022} as the fundamental module of our query component. 
As highlighted with different colors in \autoref{fig:overview} (a), MoED consists of four modules, including Bi-directional Self Attention (BiSA), Causal Self Attention (Causal SA), Cross Attention (CA), and Feed Forward Network (FFN). 
They can be assembled into three variants for various tasks:
\begin{itemize}
    \item \textbf{Text encoder} is the same as BERT \cite{devlin_bert_2019}, which encodes text by using a BiSA and an FFN in each layer. A \texttt{[CLS]} token is appended to the beginning of the text input to summarize it. 

    \item \textbf{Video-grounded text encoder} gathers task-relevant visual information by incorporating one additional cross attention between the BiSA and the FFN in each layer of the text encoder. 
    Thereinto, the text input (query or question) serves as the query and the flattened video embeddings serve as key and value.
    To accommodate the task requirements, a task-specific \texttt{[Encode]} token is appended to the text, and the resulting embedding of \texttt{[Encode]} contains the multimodal representation of the video-text pair. 

    \item 
    \textbf{Video-grounded text decoder} is responsible for collecting task-specific visual information to generate the desired text output. It replaces the BiSA layers of the video-grounded text encoder with Causal SA layers. The start of a sequence is identified using a \texttt{[Decode]} token, whereas the conclusion of the sequence is identified using an end-of-sequence token.
\end{itemize}
The subsequent discourse elaborates on how these three variations exhibit distinct approaches towards video-language understanding.

\textbf{Text-to-video retrieval.}
As summarized in \autoref{fig:overview} (c)(i), 
we perform text-to-video retrieval in two stages: 
recall and re-rank. 
They are completed by the text encoder and video-grounded text encoder, respectively. 
During inference, we first recall Top $Q$ videos by calculating the cosine similarity of the \texttt{[CLS]} token of video and text embeddings. 
Then, we re-rank the recalled videos by feeding the video-text query pair into the video-grounded text encoder, and use the output embedding of \texttt{[Encode]} with a fully-connected layer and sigmoid function to get the matching scores. 
The matching score and cosine similarity are added to get the final scores. 
The self-attention and feedforward share parameters between the text encoder and video-grounded text encoder.

\textbf{Video captioning.}
As summarized in \autoref{fig:overview} (c)(ii), our methods apply the video-grounded text decoder to generate captions based on video embeddings.

\textbf{Video question answering.}
As depicted in \autoref{fig:overview} (c)(iii), for \textit{open-ended QA}, our method first encodes video and question text into multimodal embeddings using the video-grounded text encoder. 
Then, we feed the multimodal embeddings into the video-grounded text decoder to generate answers. 
The encoder and decoder share parameters.
For \textit{multiple-choice QA}, 
we formulate it as a classification problem. Specifically, we concatenate the question and answer into a whole sentence, then we apply the video-grounded text encoder to encode the video and question-answer pair into multimodal embedding. We finally apply Softmax after a linear layer to get the score of the best answers.

\subsection{Training Objectives}
We detail the loss functions and training strategies.

\noindent \textbf{Text-to-video retrieval.}
We jointly train the text encoder and video-grounded text encoder. For the text encoder, we apply Video-Text Contrastive (VTC) loss, which aligns the video and text feature space by encouraging the \texttt{[CLS]} tokens of matched video-text pairs to have similar representations against the unmatched pairs.
Formally, for the $i$-th video-text pair, given their \texttt{[CLS]} embeddings, we follow CLIP \cite{radford_clip_2021} to apply a linear projection and $L_2$ normalization layer on them to obtain the hidden video vector $\mathbf{v}_i\in\mathbb{R}^d$ and text vector $\mathbf{t}_i\in\mathbb{R}^d$. 
First of all, 
to maximize the benefits of contrastive learning with a larger batch size
(which makes it more accurate theoretically) \cite{wu2023neighbor}, we maintain three memory banks to store most recent $M$ video vectors $\{\mathbf{v_m}\}_{m=1}^M$ and text vectors $\{\mathbf{t_m}\}_{m=1}^M$ from the momentum encoders, and the corresponding video/clip IDs $\{y_m\}_{m=1}^M$. Then we calculate the text-to-video contrastive loss $\mathcal{L}_{\textmd{t2v}}$ and video-to-text loss $\mathcal{L}_{\textmd{v2t}}$ as:
\begin{empheq}[left=\empheqlbrace]{align}
    \mathcal{L}_{\textmd{t2v}} (\mathbf{t}_i) &= 
    -\sum_{k\in\mathcal{P}(i)}
    {
        \textmd{log} \frac{\textmd{exp}(\mathbf{t}_i^T\mathbf{v}_k/\tau)}{\sum_{m=1}^M{\textmd{exp}(\mathbf{t}_i^T\mathbf{v}_m/\tau)}}
    }, \label{eq:tvl} \\
    \mathcal{L}_{\textmd{v2t}} (\mathbf{v}_i) &= 
    -\sum_{k\in\mathcal{P}(i)}
    {
        \textmd{log} \frac{\textmd{exp}(\mathbf{v}_i^T\mathbf{t}_k/\tau)}{\sum_{m=1}^M{\textmd{exp}(\mathbf{v}_i^T\mathbf{t}_m/\tau)}}
    }, \label{eq:vtl}
\end{empheq}
where $\mathcal{P}(i)=\{k|k\in M,y_k=y_i\}$ is the positive sample set, and $\tau$ is the learnable temperature parameter. Finally we combine the above two losses for the VTC loss $\mathcal{L}_{\textmd{VTC}}$:
\begin{equation}
    \mathcal{L}_{\textmd{VTC}} = \frac{1}{2} \left ( \mathcal{L}_{\textmd{t2v}} + \mathcal{L}_{\textmd{v2t}} \right ).
\end{equation}
Note that to compensate for potential false negatives in the momentum encoder, we apply the momentum distillation strategy in ALBEF \cite{li_albef_2021} to generate soft labels.

For the video-grounded text encoder, we apply Video-Text Matching (VTM) loss, which aims to learn a video-text multimodal representation that captures the fine-grained alignment between video and language. VTM corresponds to a binary classification task, where the model uses a VTM head (a linear layer) to predict whether a video-text pair is positive (matched) or negative (unmatched) given its multimodal feature of \texttt{[Encode]} token. Formally, for the $i$-th video-text pair, we first calculate their positive matching score $p_i^+\in\mathbb{R}$. Then we randomly sample a video/text to replace the matched video/text to get the negative matching score $p_i^{-v}$/$p_i^{-t}\in\mathbb{R}$. Finally, we calculate the video-text matching loss $\mathcal{L}_{\textmd{VTM}}$:
\begin{equation}
    \mathcal{L}_{\textmd{VTM}} = - \left [
        log(p_i^+) + log(1-p_i^{-v}) + log(1-p_i^{-t})
    \right ].
\end{equation}
Note that, to make the VTM loss more informative, we sample the negative samples using the hard negative mining strategy \cite{li_albef_2021}. Specifically, we use the contrastive similarity distribution from \autoref{eq:tvl} and \autoref{eq:vtl} to sample hard negatives, where similar samples have a higher chance to be sampled. We sum $\mathcal{L}_{\textmd{VTM}}$ and $\mathcal{L}_{\textmd{VTC}}$ to get the final loss.

\noindent \textbf{Video Captioning.} 
During training, we apply Language Modeling (LM) Loss for the decoder, which optimizes a cross-entropy loss that trains the model to maximize the likelihood of the text in an auto-regressive manner. Formally, for each video-text pair $(v, t)$:
\begin{equation}
    \mathcal{L}_{\textmd{LM}} = - \sum_{l=1}^L{
        \textmd{log} \left (
            P(t^l|t^{<l}, v)
        \right )
    },
\end{equation}
where $L$ is the total length of the sentence. We apply a label smoothing of 0.1 when computing the loss. Compared to the masked language modeling loss that has been widely used for video-language pretraining, LM enables the model with the generalization capability to convert visual information into coherent captions.

\noindent \textbf{Video Question Answering.} 
\textit{Open-ended QA} adopts LM loss, while \textit{multiple-choice QA} employs VTM loss. Different from that in text-to-video retrieval, the negative samples come from false question-answer pairs instead of sampling.

\section{experiments}

\subsection{Datasets}
To assess the efficacy of video-language models, we conducted experiments on three distinct tasks:
text-to-video retrieval, video caption, and video question answering. Each task was evaluated on its corresponding dataset.

\textbf{Text-to-Video Retrieval}. 
(i) \textbf{MSRVTT} \cite{xu2016msr} contains 10K YouTube videos with 200K descriptions, which is split into 9K videos for training and 1K videos for test. (ii) \textbf{DiDemo} \cite{anne2017localizing} contains 10K Flickr videos with 40K sentences, where the test set contains 1,000 videos. We follow the standard setting to concatenate all descriptions in a video as a single query, and further evaluate paragraph-to-video retrieval. (iii) \textbf{ActivityNet Captions} \cite{krishna2017dense} contains 20K YouTube videos annotated with 100K sentences. We follow the paragraph-to-video retrieval setting to train models on 10K videos and report results on the val1 set with 4.9K videos.

\textbf{Video Caption}. (i) \textbf{MSRVTT} \cite{xu2016msr} consists of 10K open-domain video clips, and each clip has 20 ground-truth captions. We use the standard captioning split, which has 6.5K training, 500 validation, and 2.9K testing videos. (ii) \textbf{MSVD} \cite{chen2011collecting} contains 1,970 videos from YouTube with 80K descriptions, which is split into 1200, 100 and 670 videos for training, validation and testing, respectively.

\textbf{Video Question Answering}. 
(i) \textbf{NeXt-QA} \cite{xiao2021next} contains about 47.7K manually annotated questions for multi-choice QA collected from 5.4K videos. There are three types of questions:  descriptive, temporal, and causal.
(ii) \textbf{MSRVTT QA} \cite{xu2017video} contain 50K QA pairs that focus on the description of video elements. For both two datasets, we use the val split for model selection, and report the final results on the test split.

\subsection{Experimental Settings}

\subsubsection{Experimental Details}
We present the key implementation details shared across all tasks.
Our models were trained on 8 GPUs. 
On the model architecture, we used ViT-B/16 \cite{dosovitskiy_vit_2021} as the video backbone, and BERT-base \cite{devlin_bert_2019} as the backbone of the query component. The number of layers $L$ was 12. Their parameters were initialized from BLIP-base \cite{li_blip_2022} without extra image-text datasets. 
On optimization techniques, we used an AdamW optimizer with a weight decay of 0.04. 
The learning rate was scheduled with a linear warm-up with 1,000 iterations, and cosine annealing starting at 10\% of training following \cite{xiao2022radar}. 
For each frame, we took random crops of resolution 224 × 224 as inputs (thus $P_{\textmd{in}}$ is 196) and applied RandAugment. We set $P_{\textmd{out}}$ to be also 196 for all models.
For task-specific hyper-parameters, we set the momentum distillation \cite{li_albef_2021} weight in VTC to 0.4. During inference, we used beam search with a beam size of 3. 
The number of frames $F$, segments $S$, learning rate, and batch size differ among datasets (see \autoref{sec:apd_exp_detail}).

\subsubsection{Evaluation Metrics}
For text-to-video retrieval, the metrics \cite{du2023multi, qu_dime_2021} recall at rank K (\textbf{R@K}, higher is better) calculate the percentage of test samples with the correct result found in the Top-K retrieved points to the query sample.
We report the $K=1,5,10$.
For video captioning, we adopted BLEU-4 (\textbf{B-4}), ROUGE-L (\textbf{R-L}), and CIDEr (\textbf{C}) metrics \cite{nie_secap_2022} (higher is better for all). 
For video question answering, we reported accuracy (\textbf{Acc}). Additionally, we reported the accuracy of each subset in NeXt-QA \cite{xiao2021next} dataset.

\subsection{Performance Comparison}

\begin{table*}[t]

\caption{Comparison of text-to-video retrieval. "PT" stands for video-language pretraining.}
\vspace{-1em}

\begin{tabular}{l|c|cccc|cccc|cccc}
\hline
\multicolumn{1}{c|}{\multirow{2}{*}{\textbf{Model}}} & \multirow{2}{*}{\textbf{PT}} & \multicolumn{4}{c|}{\textbf{MSR-VTT}}         & \multicolumn{4}{c|}{\textbf{DiDemo}}          & \multicolumn{4}{c}{\textbf{ActivityNet-Captions}}      \\ \cline{3-14} 
\multicolumn{1}{c|}{}                                &                              & \textbf{R@1}  & \textbf{R@5}  & \textbf{R@10} & \textbf{MdR} & \textbf{R@1}  & \textbf{R@5}  & \textbf{R@10} & \textbf{MdR} & \textbf{R@1}  & \textbf{R@5}  & \textbf{R@10} & \textbf{MdR} \\ \hline
BridgeFormer \cite{ge_bridgeformer_2022}                                         & \multirow{6}{*}{\cmark}           & 37.6          & 64.8          & 75.1          & 3.0          & 37.0          & 62.2          & 73.9        &3.0          & -             & -             & -             & -       \\
OmniVL \cite{wang_omnivl_2022}                                               &                              & 47.8          & 74.2          & 83.8           & -          & 52.4         & 79.5          & 85.4          & -          & -             & -             & -            \\
HiTeA \cite{ye_hitea_2022}                                                &                              & 46.8          & 71.2          & 81.9          & -            & {\ul 56.5}    & {\ul 81.7}    & {\ul 89.7}    & -            & 49.7          & 77.1          & 86.7          & -            \\
mPLUG2-B \cite{xu_mplug-2_2023}                                              &                              & 48.3          & 75.0          & 83.2          & -            & 52.3          & 80.8          & 87.5          & -            & -             & -             & -            & -             \\
STOA-VLP \cite{zhong_stoa-vlp_2023}                                             &                              & 50.1          & 75.5          & 83.8          & -            & 51.1          & 76.4          & 84.0          & -            & -             & -             & -            & -             \\
CLIP-ViP \cite{xue_clip-vip_2023}                                             &                              & \textbf{54.2} & \textbf{77.2} & \textbf{84.8} & 1.0  & 50.5          & 78.4          & 87.1          & 1.0  & {\ul 53.4} & {\ul 81.4} & {\ul 90.0} & 1.0  \\ \hline
ClipBERT \cite{lei_clipbert_2021}                                             & \multirow{6}{*}{\xmark}           & 22.0          & 46.8          & 59.9          & 6.0          & 20.4          & 48.0          & 60.8          & 6.0          & 21.3          & 49.0          & 63.5          & 6.0          \\
X-Pool \cite{gorti_x-pool_2022}                                               &                              & 46.9          & 72.8          & 82.2          & 2.0          & -            & -            & -             & -             & -             & -             & -             & -             \\
CenterCLIP \cite{zhao_centerclip_2022}                                           &                              & 48.4          & 73.8          & 82.0          & 1.0          & -             & -             & -             & -             & 46.2          & 77.0          & 87.6          & 2.0          \\
STAN \cite{liu_stan_2023}                                                 &                              & 50.0          & 75.2          & 84.1          & 1.5          & 49.4          & 74.9          & 84.5          & 1.0          & -             & -             & -             & -             \\
Cap4Video \cite{wu_cap4video_2023}                                            &                              & 51.4          & 75.7          & 83.9          & 1.0          & 52.0          & 79.4          & 87.5          & 1.0          & -             & -             & -             & -             \\ \cline{1-1} \cline{3-14} 
Ours                                                 &                              & {\ul 53.4}    & {\ul 76.1}    & {\ul 84.4}    & 1.0   & \textbf{57.6} & \textbf{84.1} & \textbf{89.8} & 1.0  & \textbf{53.5}    & \textbf{81.4}    & \textbf{91.9}    & 1.0  \\ \hline
\end{tabular}

\label{table:ret_baseline}

\end{table*}
\begin{table}[t]

\vspace{-1em}
\caption{Comparison of video captioning.}
\vspace{-1em}

\resizebox{\linewidth}{!}{

\begin{tabular}{c|c|ccc|ccc}
\hline
\multirow{2}{*}{\textbf{Model}} & \multirow{2}{*}{\textbf{PT}} & \multicolumn{3}{c|}{\textbf{MSR-VTT}}          & \multicolumn{3}{c}{\textbf{MSVD}}              \\ \cline{3-8} 
                                &                              & \textbf{B-4}  & \textbf{R-L}  & \textbf{CIDEr} & \textbf{B-4}  & \textbf{R-L}  & \textbf{CIDEr} \\ \hline
UniVL \cite{luo_univl_2020}                           & \multirow{3}{*}{\cmark}           & 41.8          & 60.8          & 50.0           & -             & -             & -              \\
MV-GPT \cite{seo_mv-gpt_2022}                          &                              & {\ul 48.9}    & 64.0          & 60.0           & -             & -             & -              \\
STOA-VLP \cite{zhong_stoa-vlp_2023}                        &                              & 45.8          & \textbf{68.4} & 60.2           & {\ul 64.4}    & \textbf{83.9} & \textbf{131.8} \\ \hline
Clip4Cap \cite{tang_clip4caption_2021}                   & \multirow{6}{*}{\xmark}           & 46.1          & 63.7          & 57.7           & -             & -             & -              \\
HMN \cite{ye_hmn_2022}                             &                              & 43.5          & 62.7          & 51.5           & 59.2          & 75.1          & 104.0          \\
SwinBERT \cite{lin_swinbert_2022}                        &                              & 45.4          & 64.1          & 55.9           & 58.2          & 77.5          & 120.6          \\
CMVC \cite{yang_cmvc_2022}                            &                              & 48.2          & 64.8          & 58.7           & -             & -             & -              \\
TextKG \cite{gu_text-kg_2023}                         &                              & 46.6          & 64.8          & {\ul 60.8}     & 60.8          & 75.1          & 105.2          \\ \cline{1-1} \cline{3-8} 
Ours                            &                              & \textbf{49.6} & {\ul 66.1}    & \textbf{69.3}  & \textbf{66.9} & {\ul 82.8}    & {\ul 123.4}    \\ \hline
\end{tabular}

}

\label{table:cap_baseline}

\end{table}
\begin{table}[t]

\vspace{-1em}
\caption{Comparison of video question answering.}
\vspace{-1em}

\resizebox{\linewidth}{!}{

\begin{tabular}{c|c|c|cccc}
\hline
\multirow{2}{*}{\textbf{Model}} & \multirow{2}{*}{\textbf{PT}} & \textbf{MSR-VTT} & \multicolumn{4}{c}{\textbf{NExT-QA}}                             \\ \cline{3-7} 
                                &                              & \textbf{Acc}       & \textbf{Acc}  & \textbf{Acc-D} & \textbf{Acc-T} & \textbf{Acc-C} \\ \hline
MV-GPT \cite{seo_mv-gpt_2022}                          & \multirow{5}{*}{\cmark}           & 41.7               & -             & -              & -              & -              \\
Flamingo \cite{alayrac_flamingo_2022}                       &                              & \textbf{47.4}      & -             & -              & -              & -              \\
HiTeA \cite{ye_hitea_2022}                          &                              & 45.4               & {\ul 62.4}    & {\ul 75.5}     & {\ul 58.7}     & {\ul 60.6}     \\
mPLUG2-B \cite{xu_mplug-2_2023}                        &                              & {\ul 46.3}         & -             & -              & -              & -              \\
CoVGT (PT) \cite{xiao_covgt_2023}                     &                              & 40.0               & 59.7          & 68.4           & 58.0           & 58.0           \\ \hline
ATP \cite{buch_atp_2022}                            & \multirow{6}{*}{\xmark}           & -                  & 54.3          & 66.8           & 50.2           & 53.1           \\
IGV \cite{li_igv_2022}                             &                              & 38.3               & 51.3          & 59.6           & 51.7           & 48.6           \\
HQGA \cite{xiao_hqga_2022}                           &                              & 38.6               & 51.8          & 59.4           & 52.3           & 49.0           \\
CoVGT \cite{xiao_covgt_2023}                          &                              & 38.3               & 59.4          & 66.8           & 57.0           & 58.5           \\ \cline{1-1} \cline{3-7} 
Ours                            &                              & 42.1               & \textbf{63.2} & \textbf{75.6}  & \textbf{59.6}  & \textbf{61.4}  \\ \hline
\end{tabular}

}

\label{table:vqa_basline}
\vspace{-1em}

\end{table}

In this section, we compared our method with recent state-of-the-art methods. 
Given that some baselines have several variants (differences in model capacity, 
training data, 
and post-processing).
We ensured a fair comparison by selecting baselines with similar model capacity (ViT/B-16) if multiple variants were present. 
We grouped all baselines into with/without video-language pretraining methods (denoted as PT in all tables). 
For text-to-video retrieval, we did NOT use the post-processing trick Dual Softmax Loss (DSL) \cite{cheng_dsl_2021}, and selected the baseline variants in the same manner. 
This is because DSL requires one-one mapping of text-video pairs, which is not a feasible approach in real-world scenarios. 
Furthermore, for the HiTeA \cite{ye_hitea_2022} model in the NeXt-QA dataset, we executed the model once again to report its results in the test split rather than the original validation split.

\subsubsection{Text-to-Video Retrieval}

Results on text-to-video retrieval are presented in \autoref{table:ret_baseline}. Upon careful examination of the results, three key observations can be made.
\begin{itemize}
    \item Our model consistently outperforms all methods without video-language pre-training. Furthermore, our method exhibits comparable performance to pre-training based models on MSR-VTT datasets and even surpasses their performance on DiDemo and ActivityNet-Captions datasets. These findings provide compelling evidence for the superiority of our RTQ framework, particularly considering that pre-training based approaches are trained with significantly larger amounts of data. For example, the CLIP-ViP model \cite{xue_clip-vip_2023} is trained on a dataset comprising 100 million video-text pairs, and initialized from CLIP \cite{radford_clip_2021} model. 
    
    \item Our method surpasses OmniVL \cite{wang_omnivl_2022} and mPLUG \cite{xu_mplug-2_2023}, despite their inclusion of temporal modeling and query structures. This result emphasizes the criticality of our refinement module. 

    \item 
    Our method achieves the best performance in DiDemo and ActivityNet Captions datasets, and comparable performance in the MSR-VTT dataset. 
    This is because videos in the former two datasets are all untrimmed videos with longer average duration and richer content, making our method more effective as it suffers less from information redundancy and scene complexity.
\end{itemize}

\subsubsection{Video Captioning}

Outcomes on captioning are illustrated in \autoref{table:cap_baseline}. These results have led to three notable observations.
\begin{itemize}
    \item Our model is consistently superior to all non-pretraining methods, and comparable with pretraining methods, demonstrating the superiority of our RTQ framework in generation tasks. 
    \item In contrast to pretraining methods, our model exhibits superior performance in the Bleu metric but inferior results in the Rouge metric. Given that Bleu is typically indicative of precision and Rouge measures recall, a plausible explanation is that our RTQ framework possesses superior learning capabilities, resulting in lower training error on the dataset and hence, a higher degree of precision. Conversely, pretraining methods incorporate more external knowledge and thus exhibit better recall.
    \item 
    Compared to pretraining methods on CIDEr metric, our model exhibits superiority in the MSR-VTT dataset, albeit inferiority in the MSVD dataset. 
    As CIDEr metric considers both semantic similarity and diversity, it provides a more comprehensive evaluation for video captioning.
    Given the fact that MSVD has 5 times fewer videos than MSR-VTT, it is likely that additional knowledge acquired through pre-training could improve performance in the smaller MSVD dataset.
\end{itemize}


\subsubsection{Video Question Answering}
Results of video question answering are illustrated in Table~\ref{table:vqa_basline}. There are two observations.
\begin{itemize}
    \item Our model is consistently superior to all non-pretraining methods, which demonstrates the superiority of our RTQ framework in complex video language tasks.  
    \item Our model is superior to pretraining methods in multiple-choice QA (NExT-QA), while inferior in open-ended QA (MSR-VTT). We attribute the strong open-ended ability of pretraining models to the modal capacity and large pretraining corpus. 
    Specifically, Flamingo has 80B parameters in total, while ours has only around 400M. 
    Besides, Flamingo uses 2 billion image-text pairs and 27 million video-text pairs, while mPLUG-B uses 2 million video-text pairs and a large natural language corpus (WikiCorpus and common crawl) as the pre-training data.
\end{itemize}
It is noteworthy that HiTeA \cite{ye_hitea_2022} formulates MSR-VTT QA as a multiple-choice problem, where the best matching word is selected as the answer. Since all answers in MSR-VTT consist of only a single word in a limited vocabulary (1.5K), such a setting leads to an overestimation of its open-ended ability.

\subsection{Ablation Studies} \label{sec:ablation}

This section presents a series of experiments aimed at analyzing the efficacy of our model, which are reported in Table 4.

\begin{table}[t]

\caption{Ablation studies of our RTQ framework. We report accuracies on the NeXt-QA \cite{xiao2021next} testset with descriptive (D), temporal (T), and causal (C) splits.}
\vspace{-1em}


\begin{tabular}{c|cccc}
\hline
\textbf{Model}       & \textbf{Acc-C} & \textbf{Acc-D} & \textbf{Acc-T} & \textbf{Acc}   \\ \hline
Ours w/o R           & 58.01          & 71.95          & 57.55          & 60.17          \\
Ours w/o T           & 58.92          & 75.27          & 57.10          & 61.04          \\
Ours w/o Q           & 59.78          & 61.42          & 57.92          & 59.43          \\ \hline
Ours w/ divST \cite{bain_frozen_2021}       & 58.64          & 74.80          & 55.55          & 60.34          \\
Ours w/ STAN \cite{liu_stan_2023}        & 58.49          & 73.95          & 55.56          & 60.12          \\
Ours w/ Uniformer \cite{li_uniformer_2022}   & 55.23          & 67.88          & 52.56          & 56.47          \\
Ours w/ TSM \cite{lin_tsm_2019}          & 57.60          & 74.45          & 55.55          & 59.73          \\ \hline
Ours w/ token select \cite{liu_ts2-net_2022} & 55.92          & 70.08          & 56.11          & 58.32          \\ \hline
\textbf{Ours}        & \textbf{61.39} & \textbf{75.58} & \textbf{59.57} & \textbf{63.15} \\ \hline
\end{tabular}


\label{table:ablation}

\end{table}

\subsubsection{Ablation of the RTQ modules}

We first examined the distinct impacts of the R, T, and Q components on our model. We conducted three variations, namely: 1) \textbf{w/o R}, which removes the clustering layer; 2) \textbf{w/o T}, which removes the temporal modeling layer; and 3) \textbf{w/o Q}, which deletes the query layer. The experiments are conducted in the NeXt-QA \cite{xiao2021next} dataset to illustrate the differences in functionality among components. For the \textbf{w/o Q} implementation, we concatenated the question and answer candidate to form a sentence, and then formulated a matching problem between videos and question-answer pairs by computing the cosine similarity of their \texttt{[CLS]} tokens. Our observations are presented below.

\begin{itemize}
    \item The refinement module has demonstrated consistent contributions across all question types, particularly for casual ones. This outcome can be attributed to the elimination of redundant information, which enhances the model's comprehension of crucial objects, actions, and events depicted in the videos. Such improvement is beneficial for answering all categories of questions.
    \item The temporal modeling module exhibits a significant impact on temporal questions while exhibiting minimal influence on descriptive ones. This phenomenon is expected as the temporal module is intended to capture temporal cues.
    \item The query module has substantially contributed to descriptive questions, whilst demonstrating relatively minor effects on other question types. 
    This outcome can be explained by the fact that descriptive questions demand a highly detailed understanding of the video content, which is facilitated by the query module. Conversely, temporal and causal questions require a more holistic understanding of the overall video content.
\end{itemize}

\subsubsection{Ablation of other designing choices}

We then replace modules in our framework for other design choices. Both \textbf{w/ divST} and \textbf{w/ STAN} replace the message token mechanism with temporal attention \cite{bain_frozen_2021, liu_stan_2023}; \textbf{w/ Uniformer} and \textbf{w/ TSM} replaces the message token mechanism with unified transformer (temporal convolution and attention) \cite{li_uniformer_2022} and temporal shift module \cite{lin_tsm_2019}, respectively. \textbf{w/ token select} replace the clustering module with token selection module \cite{liu_ts2-net_2022}. We gain two observations.

\begin{itemize}
    \item Methods of temporal attention and temporal shifting mechanism generally harm the performance. As analyzed in \autoref{sec:t_component}, these methods necessitate the spatial consistency in the patches of input frames, which is not guaranteed in our framework.

    \item Token selection method also harms the performance. Considering that token selection method is built without any prior information. It may over-select patches, resulting fewer useful information for the query module.
\end{itemize}

\subsection{Sensitivity Analysis}

\begin{figure}[t]
    \centering
    \includegraphics[width=0.99\linewidth]{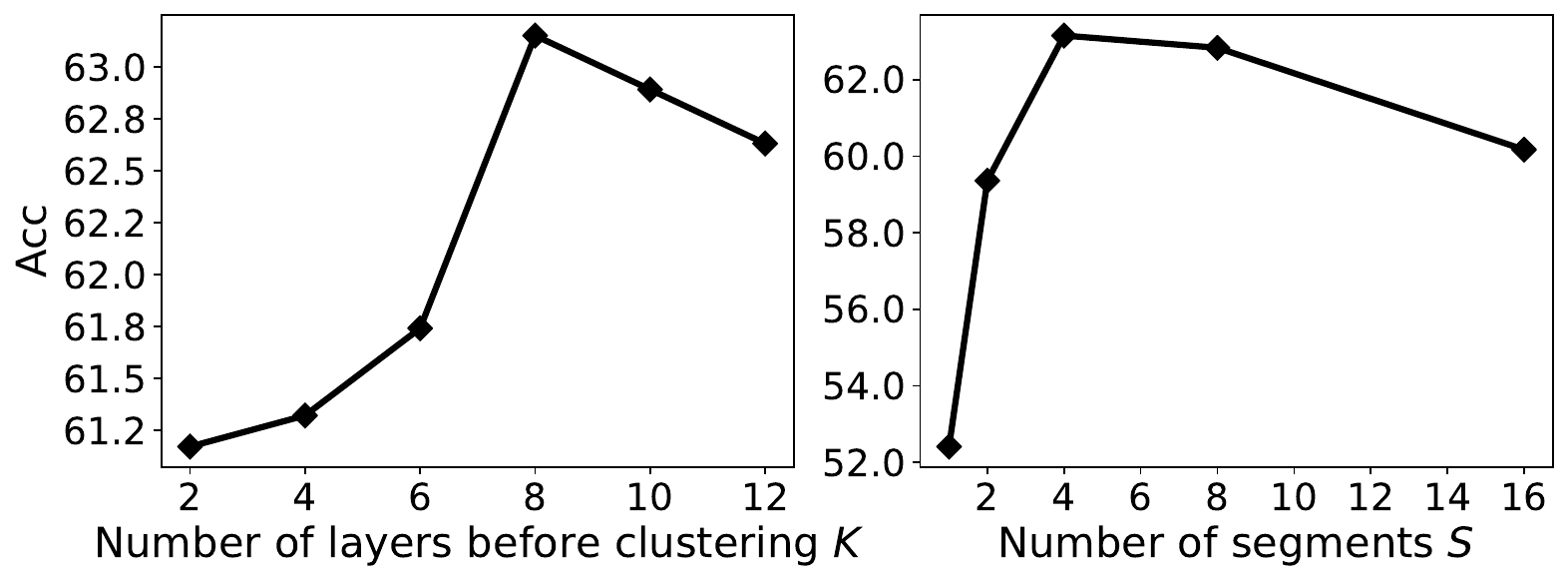}
    \vspace{-1em}
    \caption{Sensitivity analysis.}
    \vspace{-1em}
    \label{fig:sensitivity}
\end{figure}

Our study aimed to investigate the impact of hyper-parameters on the performance of our model, as illustrated in \autoref{fig:sensitivity}. Specifically, we focused on two primary hyper-parameters: the number of layers preceding the cluster layer $K$ and the number of segments of clustering $S$. We conducted experiments on the NeXt-QA dataset using 16 input frames ($F$). Our investigation leads to two key observations.

\begin{itemize}
    \item The clustering layer should be positioned within the deep layers of ViT. Specifically, the optimal number of layers before clustering is around 8. This is consistent with the design rationale outlined in Section 4. Specifically, the model should prioritize the semantic meaning of patches over their appearance in order to eliminate truly redundant patches. As deep layers contain more semantic information, they prove more effective in this regard.
    \item Secondly, we found that the optimal number of segments is approximately half of the input frames. This is a reasonable number, as reducing the number of segments increases patch purity but reduces patch integrity. Therefore, a trade-off must be struck between them.
\end{itemize}

\section{conclusion}
Our systemic analysis reveals that current video-language understanding methods focus on limited aspects of the task, and methods targeting different challenges can complement each other.
In light of this, we propose a framework integrating the refinement, temporal modeling, and query components to jointly tackle information redundancy, temporal dependency, and scene complexity, respectively.
Remarkably, 
our method achieves superior (or comparable) performance to state-of-the-art pretraining methods without requiring video-language pretraining.

To further enhance the performance of our model, there are several potential directions. 
Firstly, video-language pretraining could be used to acquire more world knowledge, which is especially helpful for open-ended QA tasks. 
Secondly, developing more effective refinement, temporal modeling, and query modules could elevate the overall performance of our approach.

\section{Acknowledgements}
This work is supported by the National Natural Science Foundation of China, No.: 62176137, and No.: 62006140; the Shandong Provincial Natural Science and Foundation, No.: ZR2020QF106.

\endgroup

\clearpage
\bibliographystyle{ACM-Reference-Format}
\bibliography{citations}


\begin{appendices}

\renewcommand\thefigure{\thesection.\arabic{figure}} 
\renewcommand\thetable{\thesection.\arabic{table}} 

\section{Experimental Details} \label{sec:apd_exp_detail}

\setcounter{table}{0} 

The number of frames $F$, segments $S$, top $Q$ in the recall stage, learning rate, and batch size differ among datasets. We will detail them in \autoref{table:t2v_details} to \autoref{table:vqa_details}.

\begin{table}[h]

\caption{Experimental Details of text-to-video retrieval.}

\begin{tabular}{c|cccccc}
\hline
\textbf{Dataset}     & $F$ & $S$ & lr & bs & epoch & $Q$ \\ \hline
MSR-VTT              & 12                        & 6                           & 2e-6                   & 64                  & 6                    & 128            \\
DiDemo               & 12                        & 6                           & 2e-5                   & 128                 & 10                   & 128            \\
ActivityNet-Captions & 32                        & 16                          & 2e-5                   & 64                  & 10                   & 512            \\ \hline
\end{tabular}

\label{table:t2v_details}

\end{table}
\begin{table}[h]

\caption{Experimental Details of video captioning.}

\begin{tabular}{c|ccccc}
\hline
\textbf{Dataset} & $F$ & $S$ & lr   & bs  & epoch \\ \hline
MSR-VTT          & 12               & 6                  & 5e-6 & 128 & 10    \\
MSVD             & 12               & 6                  & 2e-6 & 64  & 10    \\ \hline
\end{tabular}

\label{table:caption_details}

\end{table}
\begin{table}[h]

\caption{Experimental Details of video question answering.}

\begin{tabular}{c|ccccc}
\hline
\textbf{Dataset} & $F$ & $S$ & lr   & bs  & epoch \\ \hline
MSR-VTT          & 12               & 6                  & 2e-5 & 128 & 6     \\
NeXt-QA          & 16               & 4                  & 1e-5 & 64  & 6     \\ \hline
\end{tabular}

\label{table:vqa_details}

\end{table}

\section{Additional Illustration}

\setcounter{figure}{0} 

\begin{figure}[h]
    \centering
    \includegraphics[width=0.4\linewidth]{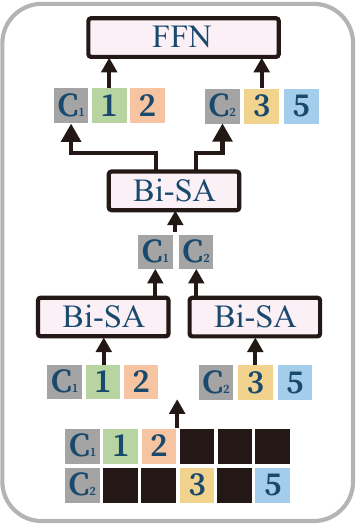}
    \caption{Illustration of the Temporal Modeling component.}
    \label{fig:apd_t_module}
\end{figure}

We illustrate the temporal modeling module of our model in \autoref{fig:apd_t_module}.

\section{Additional explanation on clustering}
On the logical relationship between the clustering results and three challenges:
The logical relationship stems from the fact that each method within the clustering graph addresses a particular challenge discussed in the introduction (analyzed in Section 2.2). Consequently, by clustering their prediction results, we can identify shared advantages and disadvantages among these methods.



\end{appendices}

\end{document}


\begingroup
\hyphenpenalty 9000
\exhyphenpenalty 9000

\title{Supplementary Materials for RTQ: Rethinking Video-language Understanding Based on Image-text Model}
\maketitle

\section{Experimental Details}

The number of frames $F$, segments $S$, top $Q$ in the recall stage, learning rate, and batch size differ among datasets. We will detail them in \autoref{table:t2v_details} to \autoref{table:vqa_details}.

\begin{table}[h]

\caption{Experimental Details of text-to-video retrieval.}

\begin{tabular}{c|cccccc}
\hline
\textbf{Dataset}     & $F$ & $S$ & lr & bs & epoch & $Q$ \\ \hline
MSR-VTT              & 12                        & 6                           & 2e-6                   & 64                  & 6                    & 128            \\
DiDemo               & 12                        & 6                           & 2e-5                   & 128                 & 10                   & 128            \\
ActivityNet-Captions & 32                        & 16                          & 2e-5                   & 64                  & 10                   & 512            \\ \hline
\end{tabular}

\label{table:t2v_details}

\end{table}
\begin{table}[h]

\caption{Experimental Details of video captioning.}

\begin{tabular}{c|ccccc}
\hline
\textbf{Dataset} & $F$ & $S$ & lr   & bs  & epoch \\ \hline
MSR-VTT          & 12               & 6                  & 5e-6 & 128 & 10    \\
MSVD             & 12               & 6                  & 2e-6 & 64  & 10    \\ \hline
\end{tabular}

\label{table:caption_details}

\end{table}
\begin{table}[h]

\caption{Experimental Details of video question answering.}

\begin{tabular}{c|ccccc}
\hline
\textbf{Dataset} & $F$ & $S$ & lr   & bs  & epoch \\ \hline
MSR-VTT          & 12               & 6                  & 2e-5 & 128 & 6     \\
NeXt-QA          & 16               & 4                  & 1e-5 & 64  & 6     \\ \hline
\end{tabular}

\label{table:vqa_details}

\end{table}

\TODO{visualized illustration of the Temporal Modeling component}

\section{Change after review}

We incorporated two suggestions from Reviewer UCAX:
\begin{itemize}
    \item \TODO{logical relationship between the clustering results and three challenges}
    \item Add the description of $L$ in Sec 4.1.
\end{itemize}

We incorporated two suggestions from Reviewer i7xS:
\begin{itemize}
    \item \TODO{Use "information redundancy" instead of “Information sparsity”}
    \item Fix un-standardized data format in Table 1.
\end{itemize}

We incorporated one suggestion from Reviewer LqSz:
\begin{itemize}
    \item Fix the typo in the abstract: absence of video-language pertaining -> absence of video-language pre-training
\end{itemize}